\documentclass[11pt]{article}
\usepackage[preprint]{acl}

\usepackage{microtype}
\usepackage{graphicx}
\usepackage{booktabs}
\usepackage{array}
\usepackage{amsmath}
\usepackage{multirow}
\usepackage{xcolor}
\usepackage{fontspec}
\newfontfamily\bnfont{NotoSansBengali}[
  Path=./,
  Extension=.ttf,
  UprightFont=*-Regular,
  Script=Bengali,
  Scale=0.98
]
\newcommand{\bn}[1]{{\bnfont #1}}
\usepackage{enumitem}

\definecolor{codebg}{HTML}{F5F3EE}
\newcommand{\code}[1]{\texttt{\small #1}}

\title{KhatianDoc: A Human-Verified Benchmark Diagnosing Multimodal LLM Failure on Bengali Legal Land Records}

\author{
  {\normalfont\large Tasmiad Hasan \quad Arafat Zaman Ratul \quad Sarker Sadman Saalim} \\
  {\normalfont\large S.M.\,Shah Nawaz Hossain \quad Khan Raiyan Ibne Reza \quad Sumaiya Tabassum Nimi} \\
  {\normalfont\normalsize North South University, Dhaka, Bangladesh} \\
  {\normalfont\small \{tasmiad.hasan, arafat.ratul, sarker.saalim, nawaz.hossain, raiyan.reza, sumaiya.nimi\}@northsouth.edu}
}

\begin{document}
\maketitle

\begin{abstract}
Land ownership in Bangladesh is recorded in \emph{Ana-Ganda-Kora-Kranti-Til}, a base-16 positional fraction system with dedicated Unicode glyphs, no mainstream font, and no coverage in any OCR pipeline or tokenizer. The handwritten records that carry these fractions, RS Khatians, are the authoritative title record for millions of parcels and a frequent subject of civil litigation, yet no benchmark has asked whether a machine can read one. We introduce \textbf{KhatianDoc}, a four-task benchmark built from 107 real RS Khatian records from the Vumi (land) Office of Munshiganj, Bangladesh: symbol recognition, base-16-to-decimal conversion, structured field extraction, and legal document question answering over 1,634 QA pairs. Ground truth was transcribed by hand, verified by a land-law practitioner to full agreement, and anonymized through positional tokens that keep the referential distinctions multi-hop questions depend on. We evaluate six multimodal LLMs (8B to 72B+, open and closed) under a fixed zero-shot protocol. Five QA categories, 39.3\% of our stratified set, return zero correct answers from \emph{every} model; on the arithmetic task, every model that emits a number does worse than a constant-mean baseline, with exact- and near-match scores coinciding: decorrelation, not approximation. Auditing our own metrics surfaced two artifacts in opposite directions: we correct a refusal-scoring bug and report the fixed scores beside the originals, and flag an inflated metadata metric as an upper bound. KhatianDoc documents not a performance gap but the absence of a capability, with verified ground truth for future systems. Code and data, with a redacted image release, are publicly available.
\end{abstract}

\vspace{4pt}
\begin{center}
\small \textbf{Dataset:} \url{https://huggingface.co/datasets/RaiyanKhaan/KhatianDoc}
\end{center}
\vspace{2pt}

\vspace{2pt}
\noindent\textbf{Highlights:}
\begin{itemize}[nosep,leftmargin=1.25em,topsep=1pt]
\item First benchmark for Bengali RS Khatian records and Ana-Ganda base-16 fractions.
\item Six multimodal LLMs score zero on 39.3\% of legal QA: a hard reasoning floor.
\item Every model underperforms a constant-mean baseline on base-16 arithmetic.
\item A metric self-audit fixes a refusal-scoring bug and flags an inflated metric.
\item Positional-token anonymization with a redacted public image release.
\end{itemize}

\begin{figure*}[t]
  \centering
  \includegraphics[width=0.485\textwidth]{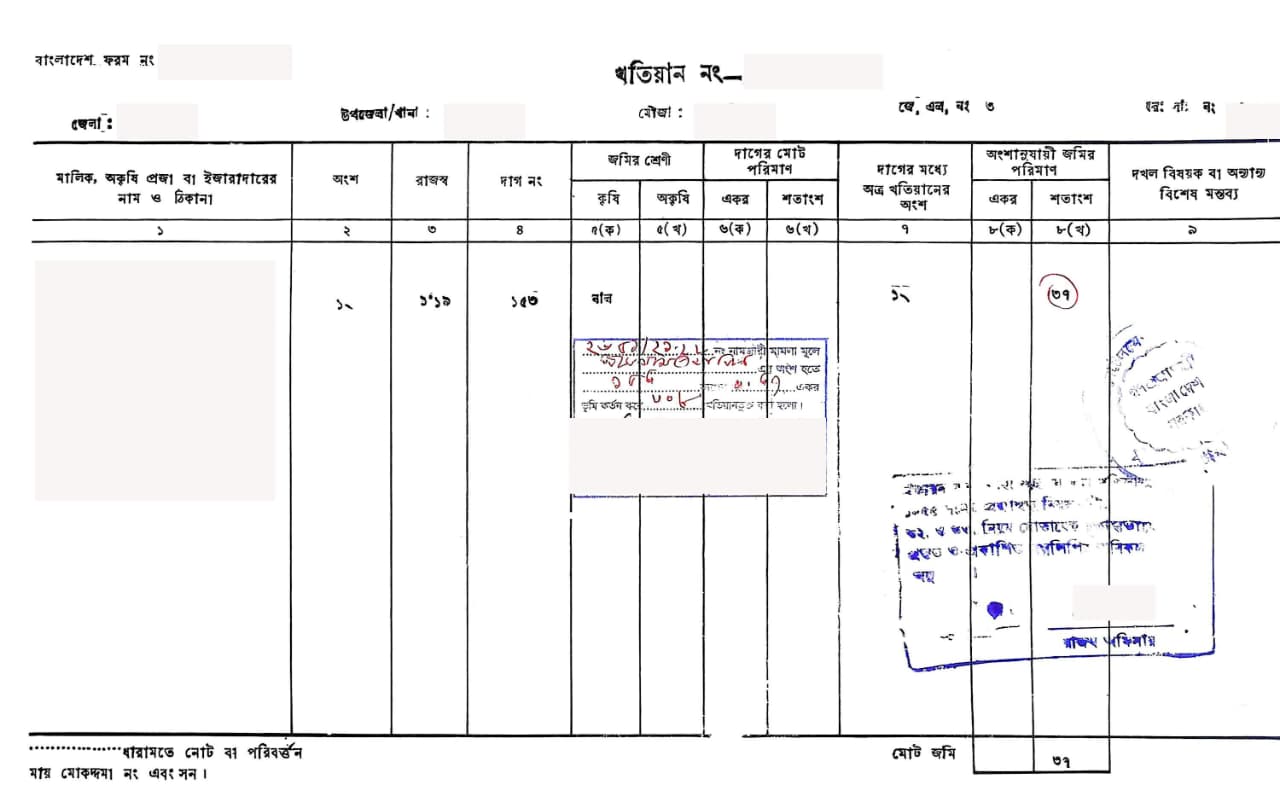}\hfill
  \includegraphics[width=0.485\textwidth]{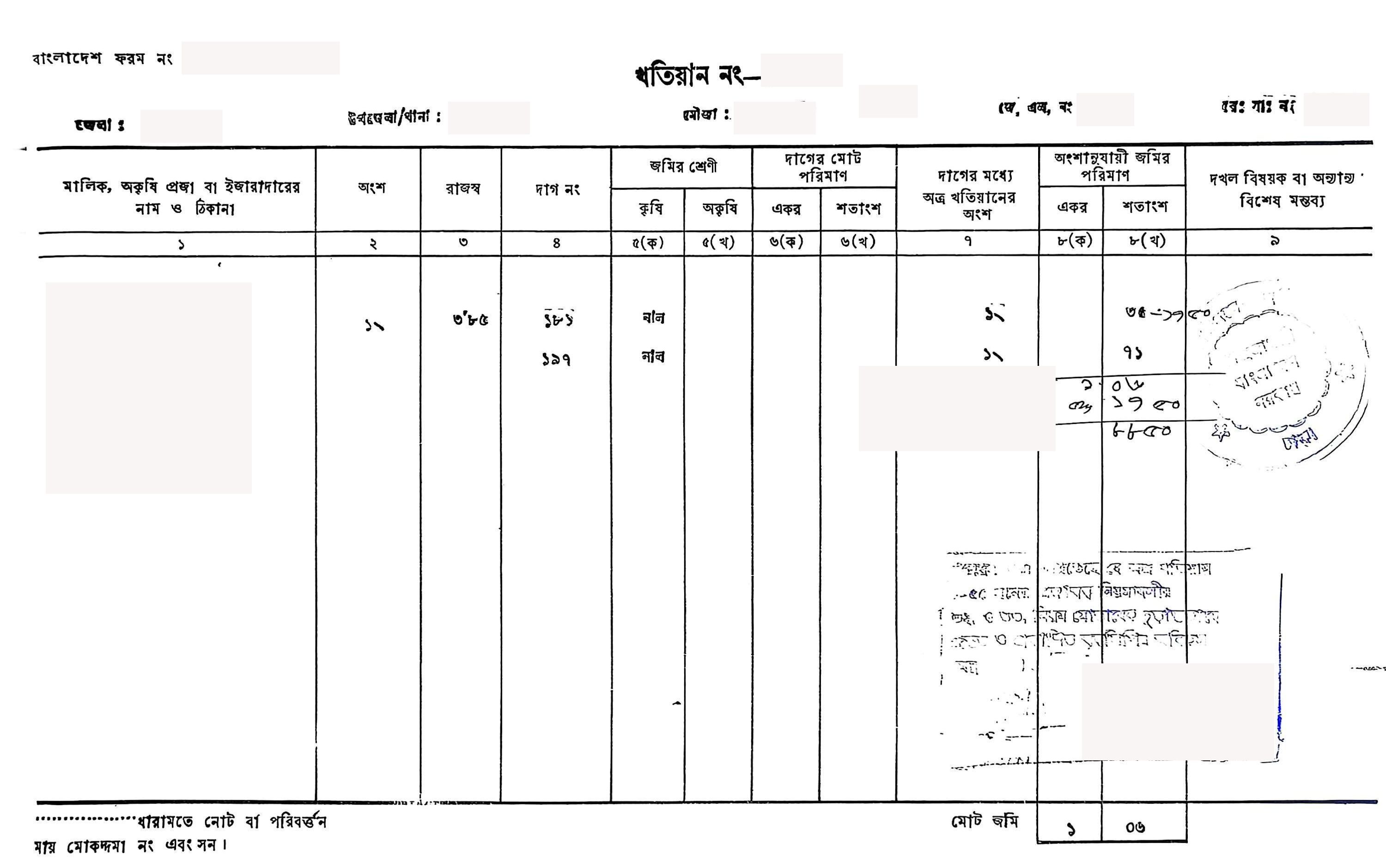}
  \caption{Two records from the \textbf{public KhatianDoc release}. Every document is a full-page scan of the same standardized RS form completed by hand. Identity-bearing regions are masked with opaque boxes before release: the owner name and address (column~1), form and khatian numbers, district / upazila / mouza headers, and the officer's seal and signature at the bottom. The fields the \emph{core} tasks score (share column \bn{অংশ}, revenue \bn{রাজস্ব}, plot number \bn{দাগ নং}, land class \bn{জমির শ্রেণী}, and the Ana-Ganda area entries) sit outside every masked region and are left intact. The Task~3 metadata header fields that fall under a mask (khatian and form number, district, upazila, mouza) are handled by a separate public-release scoring protocol, defined in \S\ref{sec:release}.}
  \label{fig:release}
\end{figure*}

\section{Introduction}

Land ownership in Bangladesh is defined, in law, by \emph{Ana-Ganda-Kora-Kranti-Til}, a base-16 positional fraction system written by hand into government survey records called RS Khatians. These records carry direct legal force in property disputes \citep{worldbank2024land}. They are the authoritative record of title for millions of parcels, and yet they sit almost entirely outside modern digital infrastructure. The Ana-Ganda glyphs have no standard font support, no OCR coverage, and no place in any NLP tokenizer vocabulary we are aware of. The legal-NLP community has built strong systems for court judgments and contracts \citep{chalkidis2022lexglue,bhattacharya2021ildc}, but nothing that has had to cope with all of what a Khatian throws at a reader at once: a degraded handwritten table, a non-decimal positional numeral system, and privacy-sensitive questions about who owns what.

We built KhatianDoc to measure that gap concretely. It draws on 107 real RS Khatian records obtained through the Vumi Office of Munshiganj district and organizes them into four tasks of increasing scope: symbol recognition, base-16 arithmetic, structured field extraction, and legal document QA over 1,634 question--answer pairs. Every field of ground truth was written out by hand and then checked, line by line, by a lawyer with land-law experience; the second pass reached complete agreement on both glyph identity and decimal value. When we ran six multimodal LLMs against it under one fixed zero-shot protocol, we did not find a smooth curve of partial competence. We found a floor. Five QA categories, together 39.3\% of the evaluation set, return zero correct answers from every model tested, and on the arithmetic task every model that produces a number lands \emph{above} the error of a baseline that ignores its input entirely and always guesses the dataset mean. Along the way, checking our scoring code against its own outputs surfaced two artifacts (one that inflates a metadata-matching metric through constant and null fields, one that penalizes models for refusing correctly), which we surface and report openly.

Our contributions are threefold.
\begin{itemize}[leftmargin=1.2em,itemsep=1pt,topsep=2pt]
\item \textbf{The benchmark.} KhatianDoc is, to our knowledge, the first benchmark for Bengali RS Khatian records, and it contains the first machine-checkable, legally verified encoding of the Ana-Ganda base-16 fraction system (\S\ref{sec:benchmark}).
\item \textbf{The diagnosis.} A zero-shot evaluation of six frontier multimodal LLMs documents universal failure on complex legal reasoning and sub-baseline performance on non-decimal arithmetic (\S\ref{sec:results},~\S\ref{sec:failure}).
\item \textbf{The methodology.} We report transferable lessons: a metric self-audit that corrects two evaluation artifacts, a privacy-preserving positional-token pipeline that keeps multi-hop QA answerable, and a redacted-image release policy for records that name real people (\S\ref{sec:benchmark},~\S\ref{sec:failure},~\S\ref{sec:release}).
\end{itemize}

\noindent Section~\ref{sec:related} places KhatianDoc against prior work; \S\ref{sec:benchmark} describes how it was built; \S\ref{sec:setup} fixes the protocol; and \S\ref{sec:results}--\ref{sec:failure} report results and failure analysis.

\section{Related Work}
\label{sec:related}

\paragraph{Document IE and legal DocVQA.}
Document-understanding models \citep{xu2020layoutlm,huang2022layoutlmv3,kim2022donut,mathew2021docvqa} and the corpora they are trained on \citep{jaume2019funsd,huang2019sroie,park2019cord,smock2022doclaynet} overwhelmingly assume typed, Latin-script forms. Legal-domain benchmarks \citep{chalkidis2022lexglue,hendrycks2021cuad} work over clean digital prose. None has met a document whose single most important numeric field is written in a base-16 positional script with no font and no OCR support, exactly the setting KhatianDoc's Tasks~1 and~3 probe.

\paragraph{Multilingual and low-resource legal NLP.}
ILDC \citep{bhattacharya2021ildc}, MultiEURLEX \citep{chalkidis2022multieurlex}, IndicBERT \citep{doddapaneni2023indicbert}, and BanglaBERT \citep{islam2022banglabert} all target typed, modern prose in base-10. RS Khatians are handwritten, heavily abbreviated, tabular, and carry their legally binding values in a numeral system absent from every tokenizer vocabulary we checked. The gap here is not one of language coverage but of script- and numeral-system-level invisibility.

\paragraph{Quantitative and symbolic reasoning.}
GSM8K \citep{cobbe2021gsm8k}, MATH \citep{hendrycks2021math}, MathVista \citep{lu2024mathvista}, NumGLUE \citep{gupta2022numglue}, FinQA \citep{chen2021finqa}, and LegalBench \citep{guha2023legalbench} test arithmetic exclusively in base-10. KhatianDoc's Task~2 asks for something with a different algebraic shape (a positional base-16 fraction with four nested subunits) that no reasoning benchmark we know of instantiates. The sub-baseline results in \S\ref{sec:results} suggest that it is this structural novelty, not mere difficulty, that drives the failure.

\paragraph{Privacy-preserving legal AI.}
Standard de-identification collapses every name onto a single generic tag \citep{lison2021anonymisation,sweeney2002kanonymity,li2022dp}. For multi-hop legal QA this quietly breaks the task: two parties replaced by the same \code{[NAME]} token make a share-comparison question unanswerable by construction. KhatianDoc's positional-token pipeline (\S\ref{sec:anon}) preserves the referential distinctions multi-hop reasoning needs, a design choice we have not seen in prior legal-NLP benchmarks.

\section{The KhatianDoc Benchmark}
\label{sec:benchmark}

KhatianDoc is built from real government land records and verified end to end without any automated transcription. This section covers how the documents were sourced, transcribed, anonymized, and cut into tasks. \S\ref{sec:setup} then describes how we test models against them.

\subsection{Source documents}
\label{sec:source}
The benchmark consists of 107 RS (Revisional Survey) Khatian records obtained through the Vumi (\bn{ভূমি}) Office of Munshiganj district, Bangladesh. Each record is a full-page scan of a printed government form filled in by hand, recording ownership, plot boundaries, and fractional inheritance shares for a single Mouza, Kumariya (\bn{কুমারিয়া}). Six documents run onto a second physical page; we keep those as paired scans (\code{\_p1}/\code{\_p2}) rather than stitching them, so that document boundaries match what the office actually issued. Because every record shares one issuing office, one Mouza, and one standardized form, the corpus isolates a single, well-defined administrative genre: a legal instrument in current use, not a synthetic imitation of one. Figure~\ref{fig:release} shows two examples from the release.

\subsection{Ground-truth construction}
\label{sec:gt}
Every field in KhatianDoc (metadata, table rows, and Ana-Ganda symbol strings alike) was transcribed by hand directly from the scan. No OCR system was used at any point, and this was deliberate. The Ana-Ganda Unicode range (\S\ref{sec:anaganda}) has no reliable font or OCR support, so leaning on automated transcription for the ground truth would have folded in precisely the errors the benchmark is meant to measure in the systems under test. Annotators produced a first transcription of each document; a Bangladeshi lawyer with land-law expertise then reviewed every transcription independently in a second pass, paying particular attention to the fraction strings, whose reading carries direct legal weight for the recorded share. That second pass reached complete agreement with the corrected transcriptions on both symbol identity and decimal value, and its output is the ground truth used throughout the paper. The corpus is not large, but it is small in the way a gold set is small: every value in it has been read twice by a human, once by a lawyer, and none of it was guessed by a machine.

\subsection{Privacy-preserving anonymization}
\label{sec:anon}
RS Khatians are live legal records that name real individuals: owners, their fathers or husbands, and residence prefixes. After transcription, we replaced every such field with a positional token (\code{[PERSON\_1]}, \code{[PERSON\_2]}, \dots), applied consistently across the Task~3 structured records and carried through the Task~4 question, answer, and evidence-span pipeline. We use positional rather than generic tokens on purpose: several Task~4 questions are multi-hop and turn on telling two named parties in the same document apart, and a single flat \code{[NAME]} token would erase that distinction and make part of the QA unanswerable. Structural and legal fields (Mouza, khatian number, land class, share fractions) are left untouched \emph{in the text ground truth}, since anonymizing them would delete exactly what the tasks are built to test. The released page images are handled separately: they additionally mask the header regions carrying the khatian and form numbers and the district/upazila/mouza labels, so the image release and the text release do not expose the same fields. We make this split explicit as a two-tier scoring protocol in \S\ref{sec:release}, since it is the one place where the public artifacts and the secure originals diverge.

\subsection{The Ana-Ganda-Kora-Kranti-Til system}
\label{sec:anaganda}
Bangladeshi ownership shares are recorded in a base-16 positional fraction system, Ana-Ganda-Kora-Kranti-Til (\bn{আনা}-\bn{গণ্ডা}-\bn{কড়া}-\bn{ক্রান্তি}-\bn{তিল}), not in base-10 decimals. The primary unit is the \emph{Ana}: 16~Ana make one full plot, and Ana values are written with dedicated Unicode glyphs in the range U+09F4--U+09F9. Four further subunits (Ganda, Kora, Kranti, Til) refine a share below a single Ana, to a precision no ordinary base-10 fraction expresses directly (full conversion rules and a worked example are in Appendix~\ref{app:anaganda}). This is the current legal standard for property division in the country, and it lives almost entirely outside mainstream digital typography: the glyphs have no standard font support, and no OCR pipeline we tested recognizes them at all. To our knowledge, KhatianDoc is the first dataset to encode Ana-Ganda values as machine-checkable ground truth.

\subsection{Task definitions and statistics}
\label{sec:tasks}
KhatianDoc has four tasks of widening scope, from an isolated glyph to a full document, summarized in Table~\ref{tab:tasks}.

\textbf{Task~1 (Symbol recognition)} pairs 95 grayscale crops of Ana-Ganda symbols with ground-truth Unicode transcriptions spanning 52 distinct strings. The crops are 256$\times$256 PNGs drawn from the documents (80 in-the-wild crops plus 15 canonical unit symbols), and the label set is dominated by three glyphs: the 4-Ana mark \bn{৷} (105 occurrences), the 1-Ana mark \bn{৴} (82), and the Bengali digit \bn{১} (54).

\textbf{Task~2 (Fraction-to-decimal)} reuses those same strings as text input and asks for each one's decimal value under the base-16 rule, over 53 distinct targets. Labels are constructed to line up with Task~1 exactly, so recognition and arithmetic can be scored as two independent stages over the same underlying values.

\textbf{Task~3 (Field extraction)} asks for a document-level metadata tier and a row-level tier (261 rows across 107 documents) from each full-page scan.

\textbf{Task~4 (Document QA)} derives 1,634 question--answer pairs from the Task~3 ground truth: 1,206 across six simple categories and 428 across four complex categories that require multi-step reasoning. A stratified 300-question subset (73.7\% simple, 26.3\% complex) is the standardized evaluation set used throughout the paper. Full taxonomies and example questions are in Appendix~\ref{app:tasks}.

\begin{table}[t]
\centering
\small
\begin{tabular}{@{}llll@{}}
\toprule
\textbf{Task} & \textbf{$N$} & \textbf{Key stat} & \textbf{Metric} \\
\midrule
T1 Symbol Rec.\  & 95    & 52 strings & CER, EM \\
T2 Fraction$\rightarrow$Dec.\ & 95 & 53 values & Exact, MAE \\
T3 Field Extr.\  & 107   & 261 rows   & F1, EM \\
T4 Document QA   & 1{,}634 & 300 eval & EM, ANLS \\
\bottomrule
\end{tabular}
\caption{KhatianDoc task statistics. Task~4 results use the 300-item stratified evaluation subset (73.7\% simple, 26.3\% complex).}
\label{tab:tasks}
\end{table}

With a human-verified, privacy-preserving ground truth in place across all four tasks, we now ask whether current multimodal LLMs can use it.

\section{Experimental Setup}
\label{sec:setup}
We evaluate six multimodal LLMs (Table~\ref{tab:models}, Appendix~\ref{app:models}) spanning closed commercial APIs and open weights, from 8B to over 70B parameters where disclosed. All six are reached through a single OpenRouter key with identical retry and rate-limit logic, which removes provider-specific client code as a confound. Every task runs under fixed, zero-shot, deterministic instructions: no in-context examples, no chain-of-thought elicitation. Blocked or dropped API responses are logged as empty strings rather than discarded or silently retried, so an empty prediction in our results is a recorded outcome, not a missing data point.

Generation length is capped per task: 30 tokens for Tasks~1 and~2, 500 for Task~3, and 200 for Task~4, matched to the expected output length of each. Scoring uses CER and EM for Task~1; Exact Decimal Match and MAE for Task~2; metadata EM and row-level F1 for Task~3; and EM with ANLS for Task~4. Prompts, formal metric definitions, thresholds, and infrastructure details are in Appendices~\ref{app:models}--\ref{app:prompts}.

\section{Results}
\label{sec:results}
Table~\ref{tab:primary} reports primary metrics for all six models across the four tasks. No model clears 26\% on any single metric except Task~3 metadata EM, where several score higher for a reason \S\ref{sec:audit} traces to constant and null fields rather than extraction, and individual metrics repeatedly bottom out at exactly zero.

\begin{table*}[t]
\centering
\small
\begin{tabular}{@{}lccccccc@{}}
\toprule
\textbf{Model} & \textbf{T1 CER}$\downarrow$ & \textbf{T1 EM}$\uparrow$ & \textbf{T2 Exact}$\uparrow$ & \textbf{T3 RowF1}$\uparrow$ & \textbf{T3 Meta-EM}$\uparrow$ & \textbf{T4 ANLS}$\uparrow$ & \textbf{T4 EM}$\uparrow$ \\
\midrule
Gemini 2.5 FL      & 93.29 & 0.00 & 2.11 & 5.16 & 45.79 & \textbf{21.12} & \textbf{13.33} \\
Qwen2.5-VL-72B     & 80.91 & \textbf{1.05} & \textbf{6.32} & 11.97 & 9.81 & 5.88 & 0.34 \\
Qwen3-VL-8B        & \textbf{76.89} & 0.00 & 2.11 & 17.48 & 11.99 & 19.54 & 13.00 \\
Llama 4 Scout      & 90.91 & \textbf{1.05} & 0.00 & 6.43 & 35.83 & 17.84 & 11.33 \\
Gemma 4 26B        & 84.70 & 0.00 & 5.26 & \textbf{25.84} & 30.84 & 4.68 & 0.33 \\
GPT-4o Mini        & 96.56 & 0.00 & 1.05 & 0.10 & 2.02 & 3.48 & 0.00 \\
\bottomrule
\end{tabular}
\caption{Primary metrics per model, all values \%. Best per column in bold. No model clears 26\% on any metric except Task~3 metadata EM, which \S\ref{sec:audit} shows is inflated by constant and null fields rather than extraction.}
\label{tab:primary}
\end{table*}

\subsection{A reasoning floor: five categories at zero}
\label{sec:floor}
\begin{figure*}[t]
  \centering
  \includegraphics[width=0.92\textwidth]{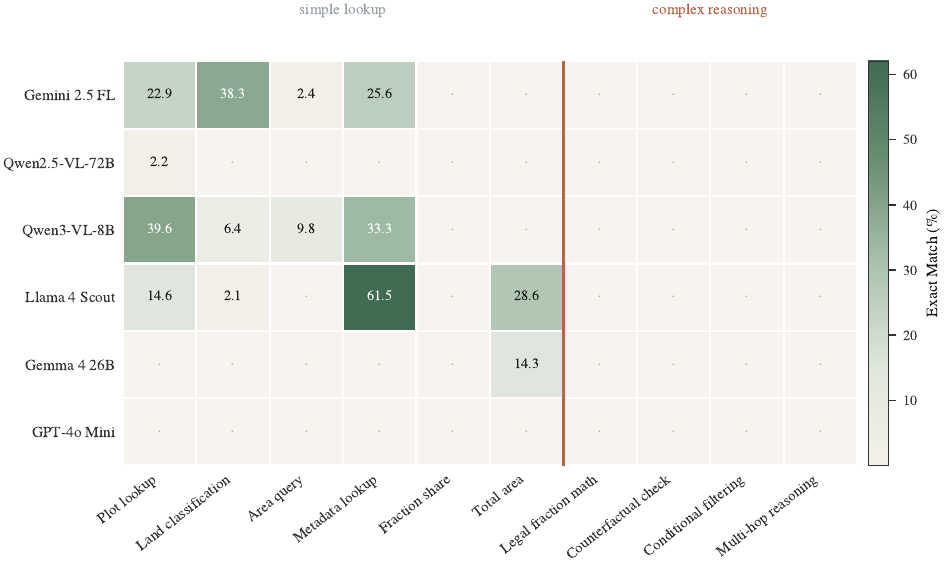}
  \caption{Task~4 Exact Match (\%) by category and model on the 300-item stratified subset. The red line separates simple lookup categories (left) from complex reasoning categories (right). Every complex category, plus \emph{fraction share}, is zero for all six models; the only non-zero cells are scattered across the simple lookups. A dot marks a true zero.}
  \label{fig:heatmap}
\end{figure*}

Figure~\ref{fig:heatmap} breaks Task~4 down by category and model over the 300-item subset, and the shape of it is hard to miss. Four categories (fraction share, legal fraction math, counterfactual check, and conditional filtering) plus multi-hop reasoning return zero correct answers from every one of the six models, with no single exception across 118 questions. This holds across the full range of scale and access we tested, from an 8B open model to 72B+ closed frontier systems. A fifth category, total area, comes close: four of the six models score zero on it and the remaining two recover only a handful of answers each, so we report it alongside rather than folding it into the zero-exception count. Together these five categories make up 39.3\% of the stratified evaluation set (Table~\ref{tab:zero}). The simple lookup categories on the left of Figure~\ref{fig:heatmap} show the only real signal in the task, and even there it is patchy and model-specific.

\begin{table}[t]
\centering
\small
\begin{tabular}{@{}lcc@{}}
\toprule
\textbf{Category} & \textbf{$n$} & \textbf{Models at 0\%} \\
\midrule
fraction\_share        & 39 & 6/6 \\
legal\_fraction\_math  & 20 & 6/6 \\
counterfactual\_check  & 20 & 6/6 \\
conditional\_filtering & 20 & 6/6 \\
multi\_hop\_reasoning  & 19 & 6/6 \\
\midrule
Total                  & 118 & 39.3\% of 300 \\
\midrule
total\_area (cf.)      & 7  & 4/6 \\
\bottomrule
\end{tabular}
\caption{Task~4 categories with zero success across all six models (top). \emph{total\_area} is shown for comparison (bottom): four of six score zero, the other two answer a few correctly.}
\label{tab:zero}
\end{table}

\subsection{Arithmetic below a context-free baseline}
\label{sec:arith}
\begin{figure}[t]
  \centering
  \includegraphics[width=\columnwidth]{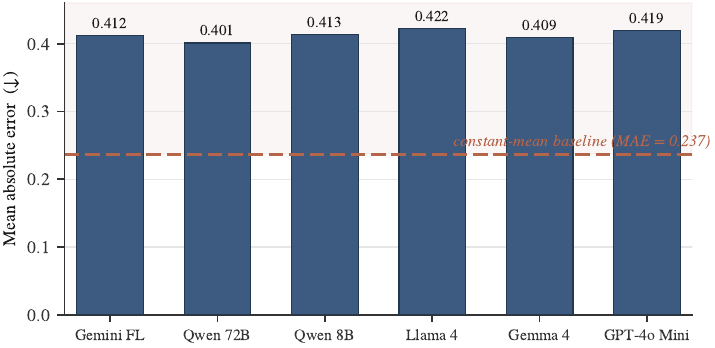}
  \caption{Task~2 mean absolute error against a constant-mean baseline (dashed, MAE $=0.237$). Every model that emits a number does worse than a predictor that ignores its input entirely.}
  \label{fig:mae}
\end{figure}

Task~2 strips out visual recognition: models get the symbol string as text, together with the base-16 conversion rule, and must return a decimal. A trivial baseline that always predicts the dataset's mean decimal value (0.3935) achieves a mean absolute error of 0.237. Figure~\ref{fig:mae} shows that every model with a defined numeric output finishes \emph{above} that error, between 0.40 and 0.42, despite being handed the conversion rule verbatim in the prompt. And for every model in Table~\ref{tab:primary}, Exact Decimal Match and the more lenient Near Match (within 0.005) return the same value, meaning no model ever lands in a close-but-not-exact band. Read together, these two facts say that model outputs are decorrelated from the input string rather than approximating it. This is not partial competence at base-16 arithmetic; it is the absence of it.

\subsection{Divergent patterns within Tasks 1 and 3}
\label{sec:divergent}
\begin{figure}[t]
  \centering
  \includegraphics[width=\columnwidth]{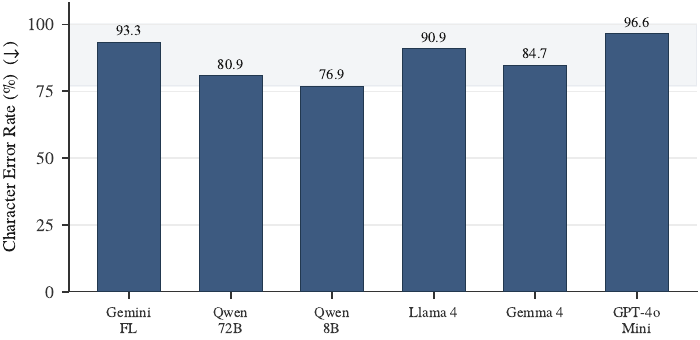}
  \caption{Task~1 Character Error Rate (\%). The band is uniformly high and roughly scale-independent, from an 8B open model to 72B+ closed systems.}
  \label{fig:cer}
\end{figure}

Task~1 CER sits in a narrow, uniformly high band (76.89 to 96.56) regardless of scale or access (Figure~\ref{fig:cer}). A failure that flat across such different systems looks more like a shared blind spot than a capability that scales with parameters. Task~3 behaves differently: row-level F1 stays low across the board (0.10 to 25.84), while metadata EM varies far more widely and, for two models, runs well above what row F1 alone would predict. Gemini 2.5 Flash Lite adds a further wrinkle from another direction: its output fails to parse against the required JSON schema on 82 of 107 documents (the worst parse rate of any model), and yet it records the best overall Task~4 performance in Table~\ref{tab:primary}. We return to both the metadata-EM pattern and this parsing tension in \S\ref{sec:audit}.

\section{Failure Analysis and Discussion}
\label{sec:failure}
Section~\ref{sec:results} established that the failure is widespread. This section shows that it is also \emph{patterned}, points to where our own scoring code shaped the picture, and pulls the patterns into a single taxonomy.

\subsection{Symbol collapse and numeral-script substitution}
\label{sec:collapse}
Task~1 asks models to transcribe glyphs they have never seen in training, and none of them respond with random noise. Each family collapses instead toward a specific, characteristic substitute. Gemini 2.5 Flash Lite reaches for the Bengali Rupee Mark (U+09F2) on 11 of 95 items; Qwen2.5-VL-72B, Qwen3-VL-8B, and Gemma 4 26B fall back to digit-slash fraction notation; GPT-4o Mini emits characters from the Brahmi Unicode range (U+11156, U+11158). Representative predictions per pattern are in Appendix~\ref{app:collapse}.

A second, script-level failure shows up in Task~3. Both the ground truth and the prompt use Bengali-script digits, yet GPT-4o Mini emits Western Arabic digits on 97 of 107 documents, against the five other scored models that stay in Bengali or mixed script on the large majority of documents (Figure~\ref{fig:script}; full breakdown in Appendix~\ref{app:script}). Because Task~3 scoring is exact-string match, a Western-digit prediction against a Bengali-script reference scores wrong regardless of whether the underlying value is right. This conflates a wrong value with a right value in a form the metric refuses to credit, a distinction we flag here because it matters for anyone reusing exact-match scoring on mixed-script data.

\begin{figure}[t]
  \centering
  \includegraphics[width=\columnwidth]{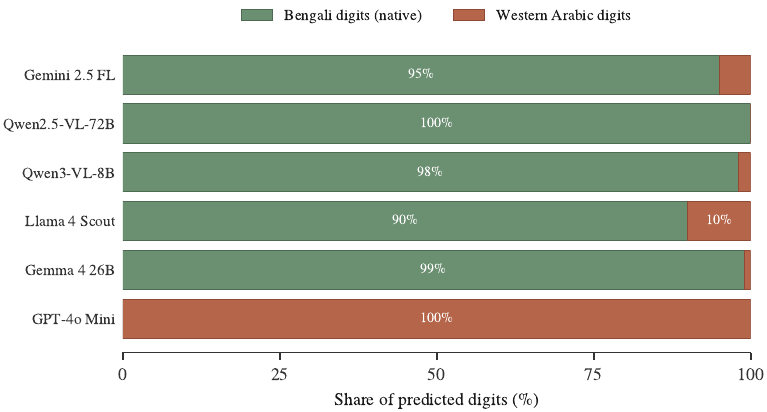}
  \caption{Numeral-script substitution on Task~3: share of predicted digits in Bengali versus Western Arabic script, for the six models scored on the task. GPT-4o Mini outputs Western digits almost exclusively, despite Bengali-script prompts and references.}
  \label{fig:script}
\end{figure}

\subsection{Two places our own metrics needed correcting}
\label{sec:audit}
Checking KhatianDoc's scoring code against its results turned up one artifact in each direction.

\paragraph{Marking a correct answer wrong.}
Task~4's prompt tells models to answer with a fixed refusal sentence whenever the requested information is absent. Several ground-truth answers for genuinely absent information instead use a short-form negation, \bn{নেই}, on its own. Gemini 2.5 Flash Lite, which follows the prompt to the letter, is scored wrong on these items despite answering exactly as instructed. A milder version of the same problem affects Llama 4 Scout, which answers \bn{খতিয়ান নং}-\bn{১৬} against a reference of \bn{১৬} alone. We report a normalized rescoring that maps the prompt's canonical refusal sentence and any ground-truth shorthand onto a single scored class; this raises Gemini 2.5 Flash Lite's Task~4 EM from 13.33\% to 21.67\% and Llama 4 Scout's from 11.33\% to 18.33\% (Table~\ref{tab:audit}). No other model's Task~4 EM changes, since only these two produced the refusal-form answers the bug mishandled.

\begin{table}[h]
\centering
\small
\setlength{\tabcolsep}{6pt}
\begin{tabular}{@{}lcc@{}}
\toprule
\textbf{Model} & \textbf{T4 EM (raw)} & \textbf{T4 EM (corrected)} \\
\midrule
Gemini 2.5 Flash Lite & 13.33 & \textbf{21.67} \\
Llama 4 Scout         & 11.33 & \textbf{18.33} \\
\bottomrule
\end{tabular}
\caption{Refusal-normalized rescoring of Task~4 EM (\%), raw values from Table~\ref{tab:primary} beside their corrected counterparts. Only these two models are affected; all other Task~4 EM values are unchanged.}
\label{tab:audit}
\end{table}

\paragraph{Marking an absent answer correct.}
Task~3 metadata EM has the opposite problem: it rewards fields that need no reading of the image. Two of its six fields (\code{district} and \code{upazila}) take a single value across the entire corpus, so a model recovers them from prior knowledge alone; and ground-truth \code{total\_area} is \code{null} for 65 of 107 documents, so any model that emits \code{null} earns a free match. A model can therefore post a sizable metadata EM by reproducing two constants and matching null, without extracting anything from the page, which is why the metadata-EM column in Table~\ref{tab:primary} runs well above every other metric. Rather than publish a single corrected number that would depend on per-field bookkeeping we do not release, we take the cleaner route: we do not treat metadata EM as a scored quantity at all. We report it raw in Table~\ref{tab:primary} as an upper bound and make row-level F1 (which credits only fields actually recovered from the image) the metric of record for Task~3.

We surface both openly because the results our claims actually rest on (the row-level extraction numbers and the core Task~4 reasoning categories) are untouched by either one.

\subsection{A failure taxonomy across models}
\label{sec:taxonomy}
Table~\ref{tab:taxonomy} collects the dominant failure pattern per model, and the groupings are distinct. The Qwen and Gemma models share a single Task~1 pattern with no Task~3 anomalies. Gemini 2.5 Flash Lite and Llama 4 Scout each show one issue of their own: a high parse-failure rate and a verbosity-driven EM gap, respectively. GPT-4o Mini stands alone in substituting Western digits for Bengali script. Taken with the metric audit, these patterns shape how Table~\ref{tab:primary} should be read: the failures are systematic, model-family-specific, and in two places partly attributable to scoring artifacts we have now surfaced.

\begin{table}[t]
\centering
\small
\setlength{\tabcolsep}{4pt}
\begin{tabular}{@{}lll@{}}
\toprule
\textbf{Model} & \textbf{T1 pattern} & \textbf{T3 note} \\
\midrule
Gemini 2.5 FL   & Rupee-mark sub.\    & parse fail 82/107 \\
Qwen2.5-VL-72B  & Digit-slash        & Bengali 83/104 \\
Qwen3-VL-8B     & Digit-slash        & Bengali 105/107 \\
Llama 4 Scout   & --                 & verbose-EM gap \\
Gemma 4 26B     & Digit-slash        & Bengali 104/104 \\
GPT-4o Mini     & Brahmi hallucin.\  & Western 97/107 \\
\bottomrule
\end{tabular}
\caption{Dominant failure pattern per model, from sampled qualitative review. A dash means no single pattern dominated. Full per-instance examples in Appendix~\ref{app:collapse}.}
\label{tab:taxonomy}
\end{table}

\section{Conclusion}
\label{sec:conclusion}
KhatianDoc makes two contributions to legal NLP. The first is a resource: 107 real Vumi-office Khatian records structured into four tasks, with ground truth built entirely by hand, checked by a legal professional, and anonymized through a positional-token pipeline that preserves the referential structure multi-hop questions need. Inside that resource, the first machine-checkable encoding of the Ana-Ganda-Kora-Kranti-Til base-16 fraction system gives the benchmark its hardest material and its clearest motivation. The second is a diagnosis: across six multimodal LLMs from 8B to 70B+ parameters, 39.3\% of the stratified evaluation set returns zero correct answers from every model, every model that emits a number underperforms a context-free baseline on the arithmetic task, and failure on both fronts is systematic and model-family-specific rather than random. Along the way we found two places where our own evaluation would have produced misleading numbers without a close audit (one inflating a metadata metric through constant and null fields, one penalizing correct refusals), and we surface both openly. A benchmark that anticipates its own critique is more useful to build on than one that does not.

\section*{Limitations}

\paragraph{Geographic and administrative scope.} All 107 documents come from a single Mouza, Kumariya, in Munshiganj district. RS Khatian forms are standardized nationally, so the document structure generalizes, but layout variation introduced by other districts, other Vumi offices, or older survey rounds is not represented in this release. Models that beat the current baseline on KhatianDoc may not carry those gains to more varied Khatian corpora without further evaluation.

\paragraph{Task~1 sample size.} The 95 symbol crops in Task~1 are drawn from that same single-Mouza corpus. Fifty-two unique strings cover a meaningful slice of the Ana-Ganda inventory, but rarer compound fractions are likely underrepresented; broader coverage would need source documents from a wider geographic base or a larger annotated corpus.

\paragraph{Zero-shot evaluation only.} We report zero-shot performance exclusively. No fine-tuning or retrieval-augmented condition is evaluated, so we cannot say how much of the observed gap is addressable with task-specific adaptation or in-context examples. We treat zero-shot as a diagnostic baseline, not a ceiling.

\paragraph{Human legal evaluation of complex QA.} The four complex Task~4 categories are scored automatically against exact or near-exact reference strings. Legal correctness and string similarity are not the same thing: a model that reasons to a legally valid answer through a valid paraphrase may be marked wrong, and a model that produces the reference string by coincidence may be marked right. Human legal-expert evaluation of complex-category predictions is left for future work.

\paragraph{OCR-augmented evaluation.} Every evaluation here runs in image-only mode, with no pre-extracted text. An OCR-augmented pipeline that passes recognized text alongside the image might shift the Task~3 and~4 failure distribution substantially. We treat image-only as the baseline condition and leave OCR-augmented results to future work, flagging their absence because the gap between the two conditions is likely large for this document type.

\section*{Ethical Considerations}

\paragraph{Data provenance.} The 107 source documents were obtained through official channels from the Vumi Office of Munshiganj district, under the standard government record-access mechanism. No document was scraped, purchased from a third party, or sourced from a private individual. These are official government survey records rather than private correspondence, though they do contain personally identifying land-ownership information, which is what motivates the handling described next.

\paragraph{Anonymization and image release.}
\label{sec:release}
Text anonymization (\S\ref{sec:anon}) removes every personal identifier from the released ground-truth files: owner names, father and husband names, and residence prefixes are all replaced by positional tokens before any file is used for annotation, training, or evaluation, and this is carried through Task~4 questions, answers, and evidence spans. No original name or residence string appears in any released text file.

The scanned images need their own answer, because a redacted transcript sitting beside an unredacted scan protects no one. We therefore define two scoring protocols and state which one every number in this paper comes from.

\emph{Secure-original protocol.} Every result reported here is computed on the original, unredacted scans, handled under access control, with all fields of all four tasks scored. These scans are not part of the open release; they are retained and made available only to vetted researchers through the same Vumi-office channel, under terms that match the sensitivity of live legal records.

\emph{Public-redacted protocol.} The open release ships \emph{redacted page images} (Figure~\ref{fig:release}): the owner name-and-address column, father and husband names, residence prefixes, form and khatian numbers, the district/upazila/mouza header labels, and the officer's seal and signature are masked with opaque boxes. Because masking removes those regions from view, the public protocol excludes the masked fields from image-based scoring rather than pretending they can still be read. Concretely, Task~1, Task~2, the row-level content of Task~3 (plot number, land class, share fractions, and the Ana-Ganda area entries), and every Task~4 question that rests on those unmasked fields are scored exactly as on the originals, since their regions are never touched. The Task~3 metadata fields that fall under a mask (khatian and form number, district, upazila, and mouza) are not scored on the public images; two of them are corpus-constants and the rest are identity-adjacent header fields, so dropping them costs the benchmark little, and \S\ref{sec:audit} already argues that metadata EM is not a load-bearing metric. Owner references stay scorable throughout, because the ground truth encodes them as positional tokens keyed to row order rather than to the masked names themselves.

The two protocols therefore measure the same four tasks, differing only in that the public one omits the masked metadata headers, a gap we state outright so that results computed on the redacted release and on the secure originals are never silently compared. In short: anonymized text and redacted images are public, original scans are access-controlled, and the paper's numbers are the secure-original ones.

\paragraph{Potential for misuse.} KhatianDoc is meant as a diagnostic benchmark for document-understanding research. The failure patterns it documents could in principle inform adversarial evaluation of legal-AI systems in South Asian jurisdictions, but those same patterns are already visible to anyone holding a standard Khatian form and an API key. The more pressing risk runs the other way: downstream land-administration tools built on models shown here to fail badly on fraction math and multi-hop legal reasoning could emit incorrect ownership outputs with real legal consequences for affected citizens. We recommend against deploying any model evaluated here on real Khatian records in an unaudited capacity.

\paragraph{Benefit to affected communities.} Land-ownership disputes are among the most common sources of civil litigation in Bangladesh, and the inaccessibility of Khatian records to digital tools is one practical obstacle to resolving them. KhatianDoc is a first step toward automated, privacy-respecting tools that could assist lawyers, surveyors, and citizens who work with these records. It is released to enable that research, not to replace the legal professionals who currently do this work.

\bibliography{custom}

\appendix

\section{Evaluated Models}
\label{app:models}
Table~\ref{tab:models} lists the six multimodal LLMs, their OpenRouter identifiers, and access type.

\begin{table}[h]
\centering
\scriptsize
\setlength{\tabcolsep}{3.5pt}
\begin{tabular}{@{}lll@{}}
\toprule
\textbf{Model} & \textbf{OpenRouter ID} & \textbf{Access} \\
\midrule
Gemini 2.5 Flash Lite & \code{google/gemini-2.5-flash-lite} & Closed \\
Qwen2.5-VL-72B & \code{qwen/qwen2.5-vl-72b-instruct} & Open \\
Qwen3-VL-8B & \code{qwen/qwen3-vl-8b-instruct} & Open \\
Llama 4 Scout & \code{meta-llama/llama-4-scout} & Open \\
Gemma 4 26B & \code{google/gemma-4-26b-a4b-it} & Open \\
GPT-4o Mini & \code{openai/gpt-4o-mini} & Closed \\
\bottomrule
\end{tabular}
\caption{Evaluated models, accessed uniformly through OpenRouter.}
\label{tab:models}
\end{table}

\section{The Ana-Ganda-Kora-Kranti-Til Numeral System}
\label{app:anaganda}
Unlike plain base-10 decimals or pure base-16 hexadecimals, the Ana-Ganda system is a mixed-radix positional fraction. Table~\ref{tab:hierarchy} gives the subunit hierarchy and Table~\ref{tab:glyphs} the dedicated Unicode glyphs.

\begin{table}[h]
\centering
\small
\begin{tabular}{@{}lll@{}}
\toprule
\textbf{Subunit} & \textbf{Relation} & \textbf{Decimal (of 1 plot)} \\
\midrule
Ana    & 16 Ana = 1 Plot   & 0.0625 \\
Ganda  & 20 Ganda = 1 Ana  & 0.003125 \\
Kora   & 4 Kora = 1 Ganda  & 0.00078125 \\
Kranti & 4 Kranti = 1 Kora & 0.0001953125 \\
Til    & 4 Til = 1 Kranti  & 0.000048828125 \\
\bottomrule
\end{tabular}
\caption{Mixed-radix subunit hierarchy of the Ana-Ganda system.}
\label{tab:hierarchy}
\end{table}

\begin{table}[h]
\centering
\small
\begin{tabular}{@{}ll@{}}
\toprule
\textbf{Codepoint} & \textbf{Glyph / description} \\
\midrule
U+09F4 & \bn{৴} Currency Numerator One \\
U+09F5 & \bn{৵} Currency Numerator Two \\
U+09F6 & \bn{৶} Currency Numerator Three \\
U+09F7 & \bn{৷} Currency Denominator Sixteen (1 Ana) \\
U+09F8 & \bn{৸} Currency Numerator Four \\
U+09F9 & \bn{৹} Currency Denominator Twenty (1 Ganda) \\
\bottomrule
\end{tabular}
\caption{Unicode codepoints used in Task~1 and Task~2. Standard Bengali numerals (\bn{০}--\bn{৯}) carry higher-order counts of these subunits.}
\label{tab:glyphs}
\end{table}

\paragraph{Worked example.} Consider a share recorded as 3 Ana, 5 Ganda, and 2 Kora. The decimal conversion sums the mixed-radix parts:
\begin{align*}
3~\text{Ana}   &= 3 \times 0.0625     = 0.1875 \\
5~\text{Ganda} &= 5 \times 0.003125   = 0.015625 \\
2~\text{Kora}  &= 2 \times 0.00078125 = 0.0015625 \\[2pt]
\text{Total}   &= 0.2046875
\end{align*}
This non-standard structure is why Task~2 injects the rule explicitly into the prompt, and why models still fail to approximate it (\S\ref{sec:arith}). Across the 95 crops the decimal values span 0.0625 to 1.19, with a mean of 0.3935 and 53 distinct targets.

\section{Dataset Statistics and Task Examples}
\label{app:tasks}

\paragraph{Corpus composition.} The 95 Task~1 crops split into compound Ana-Ganda fractions (55; 57.9\%), simple Ana symbols (35; 36.8\%), hasanta-numeric full-share marks (3; 3.2\%), a lone decimal number (1), and a lone fraction slash (1). The 107 documents yield 261 rows (mean 2.44 rows/doc, range 1--9) and 573 owner entries over 459 unique anonymized names. Sixty-three rows retain bracketed or \code{UNCLEAR} transcription markers directly from the ground truth; we keep these verbatim rather than manufacture a clean value. The Task~4 categories and their full counts are in Table~\ref{tab:t4full}.

\begin{table}[h]
\centering
\small
\begin{tabular}{@{}lr@{}}
\toprule
\textbf{Category} & \textbf{Count} \\
\midrule
plot\_lookup         & 261 \\
land\_classification & 255 \\
area\_query          & 225 \\
metadata\_lookup     & 214 \\
fraction\_share      & 212 \\
multi\_hop\_reasoning & 107 \\
legal\_fraction\_math & 107 \\
conditional\_filtering & 107 \\
counterfactual\_check & 107 \\
total\_area          & 39 \\
\midrule
Total                & 1{,}634 \\
\bottomrule
\end{tabular}
\caption{Full Task~4 QA distribution across all ten categories (six simple, four complex).}
\label{tab:t4full}
\end{table}

\paragraph{Task~3 JSON schema.} Models are instructed to output the following exact structure.
\begin{quote}\ttfamily\scriptsize
\{\\
\hspace*{1em}"metadata": \{\\
\hspace*{2em}"khatian\_no": "string", "district": "string",\\
\hspace*{2em}"upazila": "string", "mouza": "string",\\
\hspace*{2em}"rs\_no": "string", "total\_area": "string|null"\\
\hspace*{1em}\},\\
\hspace*{1em}"rows": [\{\\
\hspace*{2em}"dag\_no": "string", "land\_class": "string",\\
\hspace*{2em}"share": "string", "area\_by\_share": "string",\\
\hspace*{2em}"owners": ["[PERSON\_1]", "[PERSON\_2]"]\\
\hspace*{1em}\}]\\
\}
\end{quote}

\paragraph{Task~4 complex examples.}
\begin{itemize}[leftmargin=1.2em,itemsep=2pt,topsep=2pt]
\item \emph{Legal fraction math:} \bn{দাগ নং ১২৩ এ }\code{[PERSON\_1]}\bn{ এবং }\code{[PERSON\_2]}\bn{ এর মোট জমির পরিমাণ কত?} (What is the total land area held by Person~1 and Person~2 in Dag~123?) \emph{Requires:} row filtering, extracting two Ana-Ganda strings, converting to decimal, summing, and multiplying by total area.
\item \emph{Counterfactual check:} \bn{যদি }\code{[PERSON\_1]}\bn{ তার অংশের অর্ধেক }\code{[PERSON\_3]}\bn{ কে দান করে}, \bn{তবে }\code{[PERSON\_3]}\bn{ এর নতুন অংশ কত আনা হবে?} (If Person~1 donates half their share to Person~3, what will Person~3's new share be, in Ana?) \emph{Requires:} multi-hop extraction, base-16 division, and base-16 addition.
\end{itemize}

\section{Symbol Collapse Examples}
\label{app:collapse}
Table~\ref{tab:collapse} gives representative raw outputs for the family-specific out-of-vocabulary collapse patterns of \S\ref{sec:collapse}.

\begin{table}[h]
\centering
\small
\begin{tabular}{@{}lll@{}}
\toprule
\textbf{Model} & \textbf{Target} & \textbf{Raw output} \\
\midrule
Gemini FL & \bn{৷৴৬} & \bn{৲৲৲} (Rupee Mark) \\
Qwen 72B  & \bn{৷৴৬} & \bn{১}/\bn{২} (Digit-Slash) \\
GPT-4o    & \bn{৷৴৬} & U+11156 (Brahmi) \\
\bottomrule
\end{tabular}
\caption{Raw outputs illustrating family-specific out-of-vocabulary collapse on Task~1.}
\label{tab:collapse}
\end{table}

\section{Numeral-Script Substitution}
\label{app:script}
Table~\ref{tab:scripttab} reports the percentage of predicted digits in Bengali versus Western Arabic script for Task~3, the numeric form of Figure~\ref{fig:script}, over the six models scored on the task. GPT-4o Mini outputs Western digits almost exclusively despite Bengali-script prompts and documents.

\begin{table}[h]
\centering
\small
\begin{tabular}{@{}lcc@{}}
\toprule
\textbf{Model} & \textbf{Bengali} & \textbf{Western} \\
\midrule
Gemini FL   & 95\% & 5\% \\
Qwen 72B    & 100\% & 0\% \\
Qwen 8B     & 98\% & 2\% \\
Llama 4     & 90\% & 10\% \\
Gemma 4     & 99\% & 1\% \\
GPT-4o Mini & 0\% & 100\% \\
\bottomrule
\end{tabular}
\caption{Percentage of predicted digits in Bengali script versus Western Arabic script (Task~3).}
\label{tab:scripttab}
\end{table}

\paragraph{The metadata exploit.} As noted in \S\ref{sec:audit}, Task~3 metadata EM can be inflated without any visual extraction. Two of its fields (\code{district} and \code{upazila}) are constant across the entire 107-document corpus, and \code{total\_area} is \code{null} in the ground truth for 65 documents. A model that outputs the corpus-constant regional values and leaves the rest \code{null} therefore scores near 50\% on metadata EM without reading a single page: it collects two constants it could have memorized and a run of free null--null matches. This is why we treat the metadata-EM column as an upper bound and base extraction claims on row-level F1 instead, which credits only fields actually recovered from the image.

\paragraph{The refusal-string mismatch.} Task~4 prompts instruct: ``If absent, output exactly \bn{প্রদত্ত খতিয়ানে এই তথ্য নেই।}''. However, 14 ground-truth answers for genuinely absent fields use the shorthand \bn{নেই} alone. A model that follows the instruction perfectly (\bn{প্রদত্ত খতিয়ানে এই তথ্য নেই।}) is scored 0 against the reference \bn{নেই} under standard exact match. Normalizing both to one semantic class corrects the penalty, as detailed in \S\ref{sec:audit}.

\section{Prompt Templates}
\label{app:prompts}

\paragraph{Task~2 (fraction-to-decimal).}
\begin{quote}\small
You are an expert in Bangladeshi land measurement. Convert the following Ana-Ganda fraction string to a decimal value. Rules: 16 Ana = 1.0. 20 Ganda = 1 Ana. 4 Kora = 1 Ganda. 4 Kranti = 1 Kora. 4 Til = 1 Kranti. Output ONLY the decimal number to 10 places.\\[2pt]
Input: \code{[SYMBOL\_STRING]}\quad Output:
\end{quote}

\paragraph{Task~4 (legal QA).}
\begin{quote}\small
Answer the following question based ONLY on the provided Khatian image. Answer concisely in Bengali. If the information is not present in the document, you must output EXACTLY: \bn{প্রদত্ত খতিয়ানে এই তথ্য নেই।}\\[2pt]
Question: \code{[QUESTION\_STRING]}\quad Answer:
\end{quote}

\end{document}